# Online Reinforcement Learning-Based Dynamic Adaptive Evaluation Function for Real-Time Strategy Tasks

Weilong Yang [1], Jie Zhang [1], Xunyun Liu [1] and Yanqing Ye [2,*]

[1] Academy of Military Sciences; Beijing, 10000,P.R.China
[2] Strategic Assessments and Consultation, Academy of Military Sciences; Beijing, 10000,P.R.China
* yeyanqing09@alumni.nudt.edu.cn

**Abstract.** Effective evaluation of real-time strategy tasks requires adaptive mechanisms to cope with dynamic and unpredictable environments. This study proposes a method to improve evaluation functions for real-time responsiveness to battlefield situation changes, utilizing an online reinforcement learning-based dynamic weight adjustment mechanism within the real-time strategy game. Building on traditional static evaluation functions, the method employs gradient descent in online reinforcement learning to update weights dynamically, incorporating weight decay techniques to ensure stability. Additionally, the AdamW optimizer is integrated to adjust the learning rate and decay rate of online reinforcement learning in real time, further reducing the dependency on manual parameter tuning. Round-robin competition experiments demonstrate that this method significantly enhances the application effectiveness of the Lanchester combat model evaluation function, Simple evaluation function, and Simple Sqrt evaluation function in planning algorithms including IDABCD, IDRTMinimax, and Portfolio AI. The method achieves a notable improvement in scores, with the enhancement becoming more pronounced as the map size increases. Furthermore, the increase in evaluation function computation time induced by this method is kept below 6% for all evaluation functions and planning algorithms. The proposed dynamic adaptive evaluation function demonstrates a promising approach for real-time strategy task evaluation.

**Keywords:** online reinforcement learning, dynamic weight adjustment, real-time strategy games, evaluation functions, AdamW optimizer.

## 1 Introduction

Real-Time Strategy (RTS) tasks are renowned for their complexity and high strategic demands on human players. These games combine strategic thinking with nimble mouse operations, resulting in an intense and stimulating experience. Recently, the AI research community has increasingly focused on RTS AI research due to its numerous challenging sub-problems and stringent real-time computation constraints. The rise of esports and professional human RTS gaming has spurred interest in applying AI technologies to design, balance, and test such complex games. In RTS problems, evaluation functions play a crucial role as they determine the game AI's understanding of the current situation and its corresponding strategic choices. The accuracy and speed of these evaluations significantly impact planning performance, making faster and more accurate evaluations essential for planning methods. Traditional linear evaluation methods, though simple and computationally efficient, often fail to comprehensively capture the dynamic features of complex battlefields. To improve evaluation accuracy, researchers have introduced more complex models such as logistic regression, multi-strategy combinations, hybrid evaluation methods, and CNN-based weight learning methods. However, these approaches face challenges like strong data dependency, insufficient generalization ability, and low computational efficiency. Currently, RTS task situation assessment methods are mainly categorized into three types based on research approaches and solutions: linear methods, tree search methods, and neural network methods.

Linear methods, while simple, fail to capture complex interactions and nonlinear relationships, such as synergies between unit types and resource-time dynamics. Weight setting is cumbersome, often involving manual tuning or logistic regression, which may not account for multidimensional factors like military strength, economic development, spatial control, and player skill levels. Additionally, they struggle to adapt to dynamic game changes, leading to assessment inaccuracies. Traditional linear approaches in Real-Time Strategy (RTS) game evaluation typically employ a weighted sum of features, such as unit counts, with weights either manually tuned or learned from historical data using logistic regression. Common metrics include Life-Time Damage[1] (LTD) and unit cost, though reliance on single metrics can expose vulnerabilities[2], as demonstrated by strategies like economy rushes, which prioritize rapid economic growth but are susceptible to early attacks. Researchers have proposed more comprehensive evaluation methods to overcome these limitations. Erickson et al. (2014) [3] introduced a global evaluation function that adjusts weights using logistic regression, incorporating military, economic, spatial factors, and player skill levels. Marius et al. (2015) [4] developed an evaluation method based on Lanchester's combat laws, considering unit types and health status for more precise, dynamic assessments. These methods employ multi-dimensional evaluations and complex models to better reflect real-time battlefield situations, though their fixed weights do not adapt dynamically to changing conditions.

Tree search methods are extensively utilized for situation assessment in RTS tasks but exhibit significant limitations. They demand substantial computational resources to simulate numerous game scenarios, particularly in complex environments, resulting in high time and power requirements for reliable evaluations. The accuracy of tree search is heavily dependent on script quality, where inferior scripts lead to inaccurate simulations and unreliable assessments. Additionally, tree search is constrained by predetermined frame counts; insufficient frames fail to capture essential game dynamics, while excessive frames result in wasted computational resources. Furthermore, the complexity of tree search algorithms increases implementation and maintenance challenges, as demonstrated by Barriga et al. (2017) [5], who integrated multiple search techniques, including Alpha-Beta Considering Durations (ABCD) and Upper Confidence Bounds for Trees Considering Durations (UCTCD), to enhance precision. Crucially, tree search struggles with adaptive dynamic evaluation adjustments due to its high computational load, hindering real-time responsiveness and agility in rapidly evolving battlefield conditions.

Neural network-based situation assessment is also a research hotspot. Yang et al. (2018) [9] sought to use CNN architectures to learn generalizable, map-independent features by separating global and spatial information. Graph Neural Networks have shown promising results in modeling complex battlefield relationships[10]. Goecks et al. (2024) [11] leveraged transformer architectures for real-time combat situation understanding, achieving improved performance in development of Courses of Action. Additionally, the multi-modal deep learning approach could also robustly process various battlefield data sources for comprehensive situation assessment[12]. These developments highlight the growing potential of neural network applications in military tactical decision-making.

However, Neural network-based assessment methods also have significant limitations[13-16]. First, they require substantial computational resources and time for training and predicting, leading to slow assessment speeds, which is critical in real-time strategy games. Neural network training relies on large amounts of high-quality data, with data quality and diversity directly impacting generalization ability and assessment effectiveness. Additionally, the complexity, map size dependency, and overfitting risk of neural networks limit their general applicability.

Consequently, despite the success of existing evaluation functions in microRTS, significant limitations remain (Table 1). Manual weight setting requires extensive expert knowledge and experience, with weight selection often being subjective and suboptimal. Static weights fail to cope with real-time battlefield changes. Linear function-based assessment methods typically use static weights learned from historical data or adjusted manually [3,17]. Even CNN-based models, which adjust and optimize weights during training with large datasets, apply fixed weights during actual game assessment after training finished. These static weights cannot adapt to dynamic strategy needs during a game, leading to potential limitations in real scenarios. Yang et al. (2018) [18] proposed a dynamic hierarchical assessment network to address the limitations of traditional fixed-weight models. This method uses a hierarchical task network (HTN) planning approach, considering dynamic changes in game states and player preferences, utilizing game theory to analyze these factors and unit relationships for more accurate assessments, enhancing AI planning performance in RTS games. However, its performance heavily depends on the accuracy and comprehensiveness of domain knowledge in the hierarchical network, and it increases computation time by approximately 10%.

Moreover, high computational complexity reduces algorithm practicality. Advanced methods like CNNs improve assessment accuracy but significantly reduce computation speed, incompatible with the real-time computation needs in RTS tasks [7-9,19]. Neufeld et al. (2019) [20] proposed using evolutionary algorithms (EA) to automatically optimize evaluation function weights, aiming to enhance high-level task execution efficiency. Evolutionary algorithms adaptively optimize weights through selection and mutation operations, improving performance against multiple opponents and maps. However, they still face long evolution times and strong task sequence dependencies. Lastly, tree search methods, while evaluating win probabilities through simulated gameplay, depend on script quality and exhibit randomness [5,21-23]. Additionally, the effectiveness of limited game execution remains constrained by predetermined frame counts and time limits, failing to fully and accurately reflect real-time situation changes.

**Table 1.** Key advantages and disadvantages of existing methods.

| Method | Key Advantages | Key Disadvantages |
|---|---|---|
| Linear Methods[1-4] | Simple and intuitive<br>High computational efficiency<br>Well-adjusted Weights | Cannot capture nonlinear relationships<br>Static weights struggle to adapt to dynamic changes |
| Tree Search[5,21,22] | Deep exploration of strategy space<br>Enhanced decision quality<br>High adaptability | High computational resource consumption<br>Dependence on script quality<br>Challenging to achieve real-time responsiveness |
| Neural Network Methods[6-9] | high-dimensional feature<br>High evaluation accuracy | Long training and prediction times<br>Low computational efficiency<br>High data dependency and susceptibility to overfitting |

Addressing these limitations, our proposed method introduces a dynamic weight adjustment mechanism for traditional evaluation functions, enhanced by an online reinforcement learning framework and the AdamW optimizer. By continuously adjusting evaluation function weights in response to real-time score changes, our method ensures that the AI can promptly adapt to evolving battlefield conditions, thereby improving decision-making accuracy and effectiveness. The integration of the AdamW optimizer facilitates automatic adjustment of learning rates and decay rates, reducing the need for manual parameter tuning. This automation enhances the generalization capability and stability of the evaluation function across diverse game scenarios. Despite incorporating an online RL mechanism, our method maintains high computational efficiency, with experimental results demonstrating an average increase of less than 6% in computation time. This efficiency makes our approach suitable for real-time applications where rapid evaluations are critical.

Integrating an online reinforcement learning–based dynamic weight adjustment mechanism with traditional linear evaluation functions leverages their computational efficiency and simplicity while overcoming their static limitations, thereby providing the most effective approach for adaptive and real-time decision-making in RTS game AI. Through these innovations, our method effectively bridges the gap between the simplicity of linear evaluation functions and the adaptability of more complex methods, offering a balanced solution that enhances both performance and real-time responsiveness in RTS game AI.

## 2 Materials and Methods

### 2.1 Algorithm Process

The algorithm structure for dynamically adjusting the evaluation function weights based on online reinforcement learning, after updating the learning rate and decay rate with the AdamW optimizer, is illustrated in Fig 1. The algorithm includes three main modules: the scoring module (Module 1), the hyperparameter updating module (Module 2, AdamW optimizer), and the weight updating module (Module 3, Online Reinforcement Learning). In the RTS tasks, scoring elements can be categorized into combat units ($u$) and resources ($r$). Specifically, combat units include the main base ($MAINBASE$), barracks ($RAX$), workers ($WORKER$), light combat units ($LIGHT$), ranged combat units ($RANGE$), and heavy combat units ($HEAVY$); resources include resources already owned by the player ($R$) and in-transit resources held by worker units ($RW$).

Taking $u = MAINBASE$ as an example, $S_{MAINBASE,t}$ represents the evaluation function score for the main base at step $t$, $W_{MAINBASE,t}$ represents the scoring weight of the evaluation function score for the main base at step $t$, and $G_{MAINBASE,t}$ represents the game state information for the main base at step $t$. At the beginning of step $t$, the scoring module calculates the score based on $G_{MAINBASE,t}$, and combines it with the previous step's score $S_{MAINBASE,t-1}$ to compute the score change between the two steps $\Delta S_{MAINBASE,t}$, which is then sent to the second stage (hyperparameter updating module). In the second stage, the AdamW optimizer uses the momentum coefficients $\beta_1$ and $\beta_2$, as well as the previous step's momentum estimates $m_{MAINBASE,t-1}$ and $v_{MAINBASE,t-1}$, to calculate the momentum estimates for this step $m_{MAINBASE,t}$ and $v_{MAINBASE,t}$ based on the input $\Delta S_{MAINBASE,t}$. The bias-corrected estimates are then obtained from $m_{MAINBASE,t}$ and $v_{MAINBASE,t}$, and the adaptive learning rate ($L_t$) and adaptive decay rate ($D_t$) for step $t$ are computed. The updated hyperparameters are then fed into the third stage (weight updating module). In the third stage, the online reinforcement learning method (with hyperparameters determined from the previous stage) is used to calculate the updated weight $W_{MAINBASE,t+1}$ based on this step's weight $W_{MAINBASE,t}$ and score change $\Delta S_{MAINBASE,t}$, which will serve as the input for the next step $t + 1$.

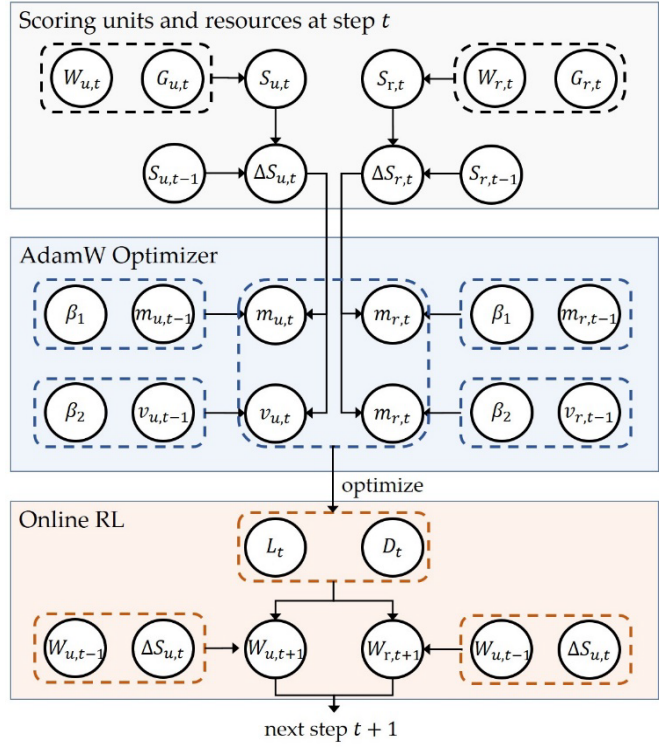

**Fig 1.** Algorithm Structure.

For dynamic weight evaluation using the online reinforcement learning algorithm, initial weights and online reinforcement parameters are required. First, initialize the unit type weights and record the initial score. When the game state needs to be evaluated, retrieve the physical game state ($G$), initialize the scores for different unit types ($S_u$), and calculate the resource score ($S_r$). Based on the difference between the current score and the previously recorded score, update the $L$, $D$ of online reinforcement learning. Then updata unit type weights ($W_u$) using online reinforcement learning. Record the latest score and weights, continuing this process until the evaluation is complete. The details of the algorithm process are illustrated in Fig 2 and Algorithm 1.

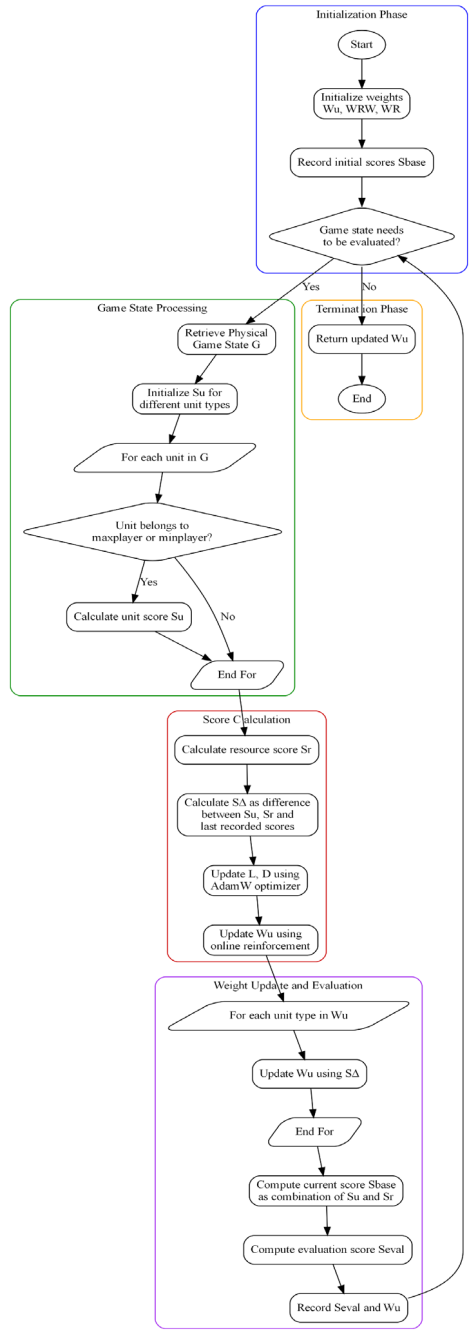

**Fig 2.** Algorithm Flow Chart. Blue: Initialization phase; Green: Game state processing; Red: Score calculation; Purple: Weight update and evaluation; Orange: Termination phase.

---

Algorithm 1 Dynamical Evaluation Function Weights Adjustment

---

**Require:** $W_u, W_{R,W}, W_R$: initial values for weights

**Require:** $\beta_1, \beta_2, \varepsilon, m_{lr}, v_{lr}, m_{dr}, v_{dr}, step$ : AdamW optimizer parameters

**Require:** $L, D$ : online reinforcement parameters

1. Initialize unit type weights in $W_u, W_{R,W}, W_R$
2. Record initial scores $S_{base}$
3. **While** game state needs to be evaluated **Do**
4. Retrieve PhysicalGameState ($G$)
5. Initialize $S_u$ for different unit types
6. **For each** unit in $G$ **Do**
7. **If** unit belongs to maxplayer or minplayer then
8. Calculate unit score $S_u$
9. **End If**
10. **End For**
11. Calculate resource score $S_r$
12. Calculate $S_\Delta$ as the difference between $S_u, S_r$ and last recorded scores
13. Update $L, D$ using AdamW optimizer: $\beta_1, \beta_2, \varepsilon, m_{lr}, v_{lr}, m_{dr}, v_{dr}, step$ with $S_\Delta$
14. Update $W_u$ using online reinforcement: $L, D$ with $S_\Delta$
15. **For** each unit type in unitWeights $W_u$ do
16. Update $W_u$ using $S_\Delta$
17. **End For**
18. Compute current score $S_{base}$ as the combination of $S_u$ and $S_r$
19. Compute evaluation score $S_{eval}$ according to $S_{base,max}$ and $S_{base,min}$
20. Record $S_{eval}$ and $W_u$
21. **End while**
22. **Return** updated $W_u$

---

In this method, $W_{unit}$ represents the weights of units, with specific weight parameters as follows: $W_{MAINBASE}$ represents the weight of the base unit; $W_{RAX}$ represents the weight of the barracks unit (which produces combat units); $W_{WORKER}$ represents the weight of the worker unit (which collects resources, constructs buildings, and can also attack); $W_{LIGHT}$ represents the weight of the light combat unit (fast and inexpensive); $W_{RANGE}$ represents the weight of the ranged combat unit (capable of long-distance attacks); $W_{HEAVY}$ represents the weight of the heavy combat unit (with high health and attack power). $S_{unit}$ represents the current round score of action units. $W_R$ represents the weight of resources already owned by the player; $W_{R,W}$ represents the weight of resources carried by worker units, and $S_{resource}$ represents the current round score of resources.

For the dynamical adjustment of $L$ and $D$ in step 13, the AdamW optimizer was used. The AdamW optimizer effectively prevents overfitting and enhances the model's generalization capability by computing parameter gradients at each time step, updating first-order and second-order moment estimates, performing bias correction, and finally applying learning rate adjustments and weight decay to update parameters. The specific process is as follows.

First, update the first-order moment estimate $m_t$ according to the score change $\Delta S_t$ at the current step $t$:

$$m_t = \beta_1 m_{t-1} + (1 - \beta_1)\Delta S_t \tag{1}$$

Second, update the second-order moment estimate $v_t$:

$$v_t = \beta_2 v_{t-1} + (1 - \beta_2)\Delta S_t^2 \tag{2}$$

where $\beta_1$ is the coefficient for the first-order moment term, $\beta_2$ is the coefficient for the second-order moment term.

Third, perform bias correction:

$$\hat{m}_t = \frac{m_t}{1 - \beta_1^t}$$

$$\hat{v}_t = \frac{v_t}{1 - \beta_2^t} \tag{3}$$

Finally, the adaptive learning rate $L_t$ and the adaptive decay rate $D_t$ for step size $t$ are computed:

$$L_t = \frac{\hat{m}_{lr,t}}{\sqrt{\hat{v}_{lr,t}} + \varepsilon}$$

$$D_t = \frac{\hat{m}_{dr,t}}{\sqrt{\hat{v}_{dr,t}} + \varepsilon} \tag{4}$$

where $m_t$ and $v_t$ represent the first moment and second moment terms, respectively. Accordingly, the bias-corrected estimates of the first moment and second moment for the learning rate are $\hat{m}_{lr,t}$ and $\hat{v}_{lr,t}$; the bias-corrected estimates of the first moment and second moment for the decay rate are $\hat{m}_{dr,t}$ and $\hat{v}_{dr,t}$; $\varepsilon$ is a small constant to prevent division by zero.

Specifically, the weight adjustments in steps 14 to 17 of the algorithm process are controlled by the learning rate ($L$) and the decay rate ($D$). After each game iteration, the adjustment of the weight $W$ for a certain unit type is calculated using the following formula:

$$W_1 = \left[W_0 + L \times \left(\frac{S_1 - S_0}{S_0}\right)\right] \times (1 - D) \tag{5}$$

Here, $W_0$ and $W_1$ are the weights before and after the update, respectively, and $S_0$ and $S_1$ are the scores before and after the update. This formula ensures that weights are updated proportionally to the relative improvement in scores (controlled by the learning rate) and also applies decay (controlled by the decay rate) to mitigate overfitting.

The evaluation function maximizes and minimizes the player's score $S_{base}$ through the following formula, providing comprehensive guidance for AI planning methods:

$$S_{eval} = 2 \times sigmoid(S_{base,max} - S_{base,min}) - 1 \qquad (6)$$

Here, $S_{eval}$ is the final score of the evaluation function, $S_{base,max}$ and $S_{base,min}$ are the scores calculated for maximizing and minimizing the player, respectively. The sigmoid function is defined as $sigmoid(x) = 1/(1 + e^{-x})$. This normalization ensures balanced evaluation between players, mitigates extreme score differences, enhances game strategy analysis, and allows comparability between scores calculated by different methods.

In our proposed algorithm, the reinforcement learning paradigm is integrated through the dynamic adjustment of evaluation function weights using an online reinforcement learning-based approach. The core idea revolves around utilizing the score change ($\Delta S$) as a feedback signal, analogous to the reward signal in traditional RL frameworks. This feedback is instrumental in guiding the adjustment of weights to better evaluate the game state in real-time.

Specifically, at each step $t$, the scoring module computes the score change $S_{MAINBASE,t}$ based on the current and previous game states. This $\Delta S$ serves as the reward signal, informing the hyperparameter updating module (Module 2) powered by the AdamW optimizer. The optimizer processes this feedback to adjust the learning rate $L_t$ and decay rate $Dt$, which are then utilized by the weight updating module (Module 3) to modify the weights $W_{MAINBASE,t+1}$. This iterative process ensures that the evaluation function remains adaptive to the dynamic battlefield conditions, thereby enhancing the AI's decision-making capabilities.

Our approach draws inspiration from traditional reinforcement learning paradigms, particularly the concepts of policy optimization and temporal difference (TD) learning[24]. By framing the weight adjustment as an optimization problem influenced by real-time feedback, we extend the RL framework to focus on the parameters of the evaluation function rather than on action selection policies directly.

Similar to how policy gradients update policy parameters based on gradient ascent on expected rewards[25], our method updates evaluation function weights based on gradient descent influenced by $\Delta S$. The use of the AdamW optimizer[26] aligns with advanced optimization techniques in deep RL, facilitating more efficient and stable convergence by adjusting learning rates dynamically based on first and second moment estimates[27]. These theoretical integrations enable our method to leverage the strengths of RL in adapting to changing environments while focusing on the specific task of evaluation function optimization.

## 2.2 Basic Evaluation Functions

Our method innovatively applies the online reinforcement learning-based dynamic weight adjustment mechanism to existing evaluation functions in microRTS based on the work of Santiago et al. (2013)[28], including the Lanchester Models (Section 2.2.1), Simple Evaluation Function (Section 2.2.2), and Simple Sqrt Evaluation Function (Section 2.2.3). The dynamic weight adjustment method for evaluation functions based on online reinforcement learning can be applied to various evaluation functions, enabling adaptive adjustments to battlefield situations by dynamically updating their weight metrics. This approach enhances their performance. This enhancement aims to improve the adaptability and accuracy of these traditional functions in RTS tasks by enabling them to respond dynamically to changing battlefield conditions. Consequently, our approach complements rather than replaces existing evaluation techniques, providing a means to enhance their performance without necessitating a complete overhaul or substitution with more complex, state-of-the-art methods.

### 2.2.1. Lanchester Models

The Lanchester combat model evaluation function, grounded in the classic Lanchester equations [29], assesses the game state by simulating the attrition process of opposing armies. The Lanchester laws consist of two primary forms: the linear law and the square law. The linear law applies to ranged weapon combat, such as archery or firearms, where each unit's attrition rate on the enemy is constant and does not vary with the increase in friendly forces.

In this study, the Lanchester evaluation function adjusts the scores based on the number of buildings (i=1 for base; i=2 for barracks), units (i=3 for workers; i=4 for light units; i=5 for ranged units; i=6 for heavy units), and resources ($res_{carried}$ for resources in transit; $res_{mined}$ for acquired resources), providing a comprehensive score for the player. The initial weights for each unit type, including base (0.129), barracks (0.231), workers (0.181), light units (1.75), ranged units (1.679), and heavy units (3.9), are determined based on the work of Santiago et al. (2013)[28]. The total score at a given moment is calculated as follows:

$$\text{score} = \sum_{i=1}^{2} \text{score}_i W_i + \left(n_a^{0.7}\right)\sum_{i=3}^{6} \text{score}_i W_i + \text{res}_{carried} \times W_{carried} + \text{res}_{mined} \times W_{mined} \quad (7)$$

where $\text{score}_i$ is the score for the corresponding unit, and $W_i$ is the weight for the corresponding unit; for instance, $\text{score}_1$ and $W_1$ represent the score and weight for the base, respectively. $n_a$ is the number of units in action. For light and heavy units, their score is the ratio of their current health to maximum health; for other units, the score is based on their health.

### 2.2.2. Simple Evaluation Function

The Classical Simple evaluation function is one of the built-in evaluation functions in microRTS, assessing the game state by evaluating players' resources and unit performance. This function combines resource evaluation with unit-based scoring and uses a sigmoid function for normalization.

The basic scoring function calculates a specific player's score in a given game state. The formula is as follows:

$$score = r \cdot R + \sum_{u \in U}(r_u \cdot R_W + \frac{U_B \cdot c_u \cdot h_u}{h_{max}}) \quad (8)$$

Here, $r$ represents the player's resource quantity; $R = 20$ is the initial weight for existing resources; $r_u$ represents the quantity of resources carried by unit $u$; $R_W = 10$ is the initial weight for resources in transit carried by units; $U_B = 40$ is the initial weight for units in action; $c_u$ represents the cost of unit $u$; $h_u$ represents the current health of unit $u$; and $h_{max}$ represents the maximum health of unit $u$.

The upper bound function calculates the highest possible score in the current game state. The formula is as follows:

$$score_{upper} = \left(R_{free} + max\left(R_{player\ 0}, R_{player\ 1}\right)\right) \cdot U_B \quad (9)$$

Here, $R_{\text{free}}$ represents the total quantity of resources held by neutral parties in the game state; $R_{\text{player 0}}$, $R_{\text{player 1}}$ represent the total resource quantities held by the two players, including resources and costs associated with their units.

### 2.2.3. Simple Sqrt Evaluation Function

The Classical Simple Sqrt evaluation function is also one of the built-in evaluation functions in microRTS. Building on the formula presented in equation (9), this function incorporates the square root of the health ratio in the basic scoring function, resulting in smoother score variations and enhanced robustness. The score calculation formula is as follows:

$$score = r \cdot R + \sum_{u \in U}(r_u \cdot R_W + U_B \cdot c_u \cdot \sqrt{\frac{h_u}{h_{max}}}) \tag{10}$$

## 2.3 Basic Planning Algorithms

The effectiveness of a situational assessment function is evidenced by the performance of the planning algorithms it guides. Here, IDABCD[30], IDRTMinimax[31], and Portfolio[32] planning methods are selected as the basic planning Algorithms.

### IDABCD Method

For challenging problems such as large-scale real-time decision-making tasks (e.g., video games and robotics), existing search algorithms face increasing runtime and memory demands. To address this, Churchill et al. (2012) [30] proposed using the Alpha-Beta search algorithm to solve adversarial real-time planning problems with durative actions and developed the ABCD (Alpha-Beta Considering Durations) algorithm and its iterative deepening version, IDABCD (Iterative Deepening Alpha-Beta Considering Durations). This algorithm has been widely used in real-time strategy games for handling small-scale combat scenarios. The algorithm is designed for adversarial real-time planning problems with durative actions and can complete search tasks on a single core within 5 milliseconds when dealing with multi-unit interactions. The ABCD algorithm reduces search nodes by considering durative actions and defining the next non-"pass" action time point. It uses depth-first search combined with dynamic pruning to handle actions with durations in each state.

In practical computation, IDABCD finds the optimal action through iterative deepening, alpha-beta pruning, and simulation games, combined with evaluation functions. First, at the Max Node, the algorithm aims to find the action that maximizes the evaluation function value, i.e., $\alpha = max(\alpha, E(G))$. If the current node's evaluation value $E(G)$ is greater than the current best value, the best action $A_{best}$ is updated. At the Min Node, the goal is to find the action that minimizes the evaluation function value, i.e., $\beta = min(\beta, E(G))$. If the current node's evaluation value $E(G)$ is less than the current best value, the best action $A_{best}$ is updated. During the search, the IDABCD algorithm uses alpha-beta pruning to reduce the search space, performing pruning when $\beta \leq \alpha$. The recursive search formula can be expressed as:

$$\begin{cases} E(G) & \text{if } d = 0 \text{ or gameover } (G) \\ max(Search(G', d-1, \alpha, \beta)) & \text{if Max Node} \\ min(Search(G', d-1, \alpha, \beta)) & \text{if Min Node} \end{cases} \tag{11}$$

where $G'$ is the new game state after applying action $A$, and $d$ is the current node's search depth. To ensure the optimal solution is found within the time budget, the algorithm employs an iterative deepening strategy, incrementally increasing the search depth until the maximum depth or time budget is exhausted. At the simulation nodes, the current game state is simulated until the maximum simulation time or game end is reached.

**IDRTMinimax**

IDRTMinimax is an iterative deepening version of the real-time Minimax algorithm [31]. Traditional Minimax algorithms are typically used in turn-based games where the AI agent has sufficient time to compute the best next move. However, in RTS games, decisions need to be made under strict time constraints, challenging the traditional Minimax algorithm. IDRTMinimax employs an iterative deepening technique to find the best decision within a limited time by gradually deepening the search depth in successive iterations, ensuring the best possible solution is found before time runs out. By utilizing the available time to search the decision tree as deeply as possible, IDRTMinimax applies the Minimax algorithm in real-time environments.

The core of the IDRTMinimax algorithm is to use a given time budget $T$ to achieve the maximum possible search depth $d$ in the decision tree through an iterative deepening strategy. First, the IDRTMinimax algorithm initializes the search depth to $d = 1$. Then, at each depth $d$, a depth-limited Minimax search is performed, combined with alpha-beta pruning. Let $S(d)$ be the search space explored at depth $d$; for each game state $s \in S(d)$, the evaluation function $E(s)$ is used for evaluation. In each step, the IDRTMinimax algorithm ($I$) monitors the elapsed time $t$. If $t < T$, the search is terminated, and the best action found at the current depth is returned. This process can be described by the following recursive formula:

$$I(s, d, \alpha, \beta, T) = \begin{cases} E(s) & \text{if } d = 0 \text{ or gameover } (G) \\ max_{a \in A(s)} I(s', d-1, \alpha, \beta, T) & \text{if Max Node} \\ min_{a \in A(s)} I(s', d-1, \alpha, \beta, T) & \text{if Min Node} \end{cases} \quad (12)$$

where $\alpha, \beta$ are the parameters for alpha-beta pruning; $A(s)$ is the set of all possible actions in state $s$; $s'$ is the resulting state after applying action $a$. In each iteration, the IDRTMinimax algorithm attempts to explore the search space $S(d)$ as deeply as possible within the remaining time and updates the best action and its corresponding evaluation score. This method ensures that within the time constraints of real-time strategy games, the algorithm can find the best possible action, balancing search depth and computational efficiency. By this approach, the algorithm guarantees identifying the best possible action within the given computation time $T$, enabling better decision-making in dynamically changing environments.

**Portfolio**

The Portfolio Greedy Search [32] (PGS) algorithm is a planning method that integrates multiple AI strategies, dynamically selecting the optimal strategy to handle different game scenarios. The algorithm first defines a script set and the initial game state, evaluating the performance of each script. Through a series of playouts, the algorithm selects the initial strategy for each unit and gradually optimizes unit strategies using a hill-climbing algorithm. For each unit, the algorithm evaluates the performance of different scripts in playouts and selects the script with the highest score as the best strategy for the current round. In the main loop, the algorithm alternates between improving the player's and the opponent's strategies within the specified response iteration times. Through alternating improvements, the algorithm gradually optimizes both players' strategies until the specified number of iterations is reached. For each script, the algorithm calculates its minimum score against all opponent strategies and selects the script with the maximum minimum score as the best strategy. By evaluating and improving strategies through multiple playouts, the Portfolio Greedy Search algorithm can find better action sequences within a limited time, enhancing search efficiency and decision quality.

## 2.4 Experiment Design

### Round-Robin Competition

This study adopts a round-robin tournament experimental approach, conducting experiments on M1 (16×16), M2 (24×24), and M3 (32×32) maps (Fig 3), with a maximum of 10,000 game rounds.

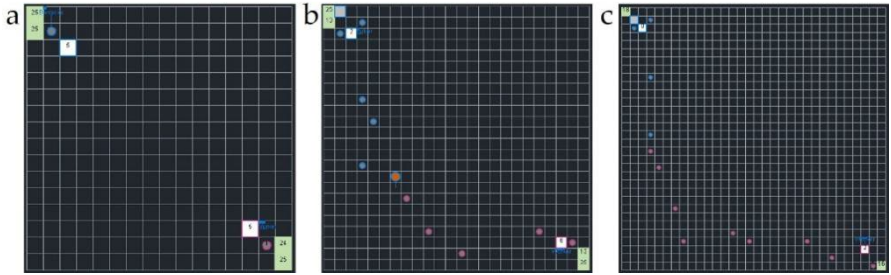

**Fig 3.** M1 (left, 16×16), M2 (center, 24×24), and M3 (right, 32×32) maps. Green squares represent neutral resources, blue-framed gray squares represent barracks built by the blue side, blue-framed orange circles represent blue side's light combat units, blue (red) framed white squares represent the blue (red) side's bases, and blue (red) dots represent blue (red) side's worker units.

In the round-robin tournament, each method competes against every other method in 40 rounds of combat (20 rounds as player 1 and 20 rounds as player 2). The evaluation schemes include the Dynamic Adaptive Lanchester Model (DL), the Dynamic Adaptive Simple Evaluation Function (DS), the Dynamic Adaptive Simple Sqrt Evaluation Function (DSQ), the Traditional Lanchester Model (L), the Classic Simple Evaluation Function (S), and the Classic Simple Sqrt Evaluation Function (SQ). The planning schemes include IDABCD, IDRTMinimax, and PortfolioAI. Consequently, when testing six evaluation functions (resulting in 15 combinations) against each other in 40 rounds of combat on three maps using three planning schemes, a total of 5,400 rounds of combat are required (15×40×3×3).

### Experimental Conditions

The experimental environment is configured with an R5-4500 processor and 32GB of memory, using Java version 22.0.1. The CPU time for each round is set to 100ms.

### Experimental Platform

The microRTS platform used in the experiments is a mature platform for researching and developing RTS algorithms. It has been widely utilized in the research community for studying RTS problems [33-37]. The platform provides a simple architecture, various game modes, and a flexible API, allowing researchers to quickly prototype and conduct experiments, and evaluate algorithm performance through competition with other AI agents.

## 3    Results

### 3.1    Experimental Scores

In the new dynamic weight adjustment method for online reinforcement learning evaluation functions based on AdamW, we selected commonly used AdamW parameters with momentum term coefficient $\beta_1 = 0.9$ and second momentum term coefficient $\beta_2 = 0.999$, and tested with an initial learning rate $L = 1 \times 10^{-4}$.

**Table 2.** Average scores of each evaluation function in different planning algorithms.

| Planning Algorithm | Dynamical Lanchester | Dynamical Simple eval. | Dynamical Simple Sqrt eval. | Lanchester | Simple eval. | Simple Sqrt eval. |
|---|---|---|---|---|---|---|
| IDABCD | 0.73 | 0.56 | 0.61 | 0.40 | 0.37 | 0.32 |
| IDRTMinmax | 0.94 | 0.42 | 0.42 | 0.67 | 0.27 | 0.27 |
| Portfolio | 0.46 | 0.74 | 0.78 | 0.40 | 0.35 | 0.28 |

Here, we compared the performance of each traditional evaluation function before and after the application of our dynamic weight adjustment method across various map sizes (M1: 16×16, M2: 24×24, M3: 32×32) and planning algorithms (IDABCD, IDRTMinmax, Portfolio AI). The results indicate that the dynamically adjusted evaluation functions consistently outperform their static counterparts, with improvements becoming more pronounced as the complexity of the game environment increases. For instance, the DL and DSQ demonstrated significant score enhancements across all map sizes and planning algorithms, as illustrated in Fig 4, Fig 5, and Fig 6. These findings are further supported by Table 2, which summarizes the average scores achieved by each evaluation function under different planning algorithms.

Specifically, Table 2 shows the average scores of each evaluating algorithm applied in IDABCD, IDRTMinmax and Portfolio. Based on the results, factors such as the specific algorithm, map size, and the characteristics of the evaluation function itself all affect the improvement effectiveness of the dynamic weight adjustment method on the evaluation function. For IDABCD, the dynamic weight adjustment method significantly improved the dynamic adaptive Simple Sqrt evaluation function and the dynamic adaptive Lanchester model across all three map sizes (M1: Fig 4,M2: Fig 5,M3: Fig 6). However, the improvement effect on the dynamic adaptive Simple evaluation function was not obvious in M1 (Fig 4), likely due to the higher randomness in smaller maps. The influence of map size on the results was more pronounced in IDRTMinimax, where the improvement of DS and DSQ evaluation functions compared to their original counterparts S and SQ became more significant as the map size increased. Conversely, in the Portfolio method, the improvement effect was not significantly affected by the map size. The DS and DSQ evaluation functions consistently showed significant improvements, while the improvement effect of DL remained consistently smaller.

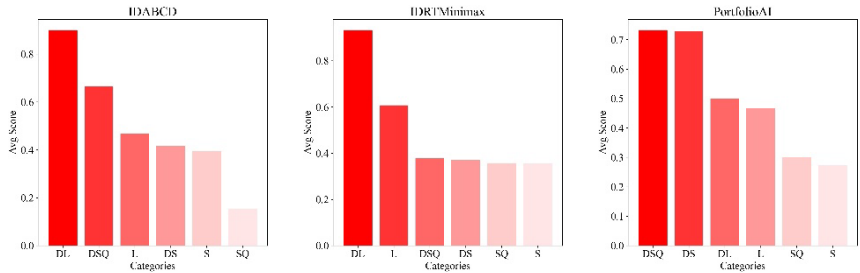

**Fig 4.** Round-robin tournament results in M1 map for Lanchester evaluation function (DL), Simple evaluation function (DS), and SimpleSqrt evaluation function (DSQ) with dynamic weight adjustment using AdamW-based online reinforcement learning; and non-dynamic Lanchester evaluation function (L), Simple evaluation function (S), and SimpleSqrt evaluation function (SQ) applied to IDABCD, IDRTMinimax, and Portfolio AI methods.

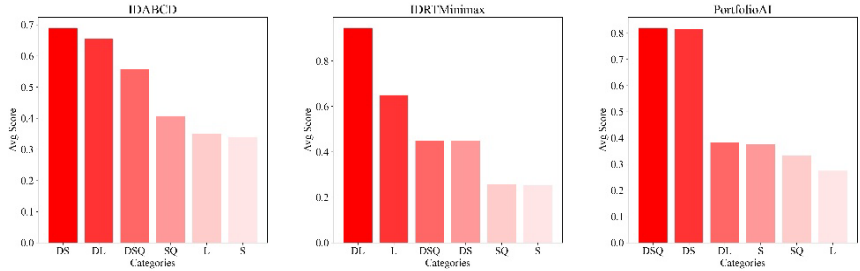

**Fig 5.** Round-robin tournament results in M2 map for Lanchester evaluation function (DL), Simple evaluation function (DS), and SimpleSqrt evaluation function (DSQ) with dynamic weight adjustment using AdamW-based online reinforcement learning; and non-dynamic Lanchester evaluation function (L), Simple evaluation function (S), and SimpleSqrt evaluation function (SQ) applied to IDABCD, IDRTMinimax, and Portfolio AI methods.

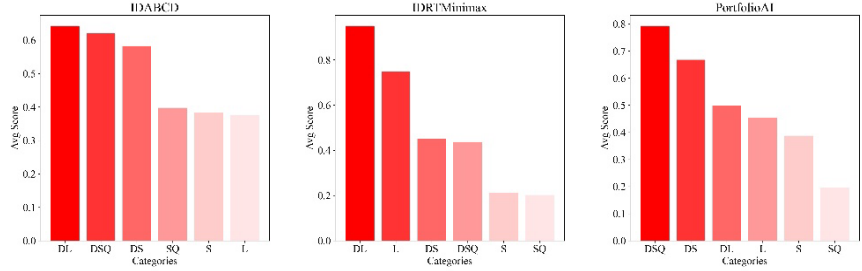

**Fig 6.** Round-robin tournament results in M3 map for Lanchester evaluation function (DL), Simple evaluation function (DS), and SimpleSqrt evaluation function (DSQ) with dynamic weight adjustment using AdamW-based online reinforcement learning; and non-dynamic Lanchester evaluation function (L), Simple evaluation function (S), and SimpleSqrt evaluation function (SQ) applied to IDABCD, IDRTMinimax, and Portfolio AI methods.

### 3.2 Computational Efficiency

Due to the nature of online reinforcement learning algorithms, which require only minimal computation at each time step to update policies and value functions without the need for large-scale data accumulation for complex global optimization, online learning methods exhibit high computational efficiency. Additionally, although AdamW includes extra computational steps, the overhead of these steps is relatively small and does not significantly increase the overall computational burden. Therefore, the dynamic weight adjustment using AdamW-based online reinforcement learning imposes a minimal computational load. According to the experimental results on M1 (Fig 7), M2 (Fig 8), and M3 (Fig 9), the additional computational time due to dynamic weight adjustment is at most 0.016ms, representing an increase of approximately 6% over the original baseline.

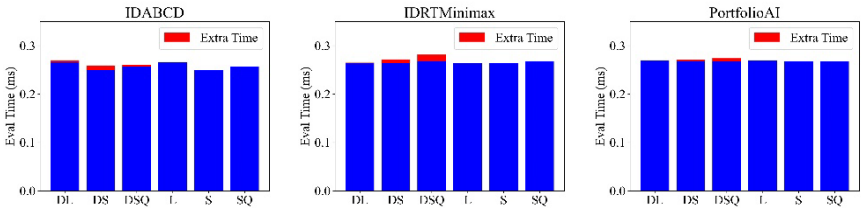

**Fig 7.** Computational efficiency in M1 for Lanchester evaluation function (DL), Simple evaluation function (DS), and SimpleSqrt evaluation function (DSQ) with dynamic weight adjustment using AdamW-based online reinforcement learning; and non-dynamic Lanchester evaluation function (L), Simple evaluation function (S), and SimpleSqrt evaluation function (SQ) applied to IDABCD, IDRTMinimax, and Portfolio AI methods.

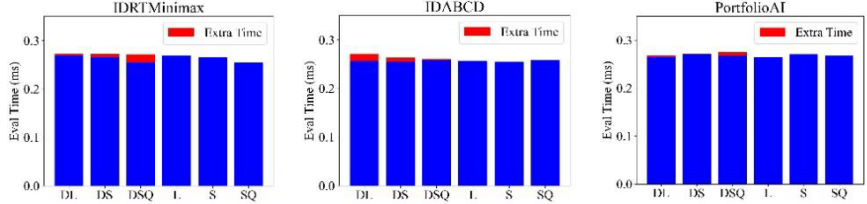

**Fig 8.** Computational efficiency in M2 for Lanchester evaluation function (DL), Simple evaluation function (DS), and SimpleSqrt evaluation function (DSQ) with dynamic weight adjustment using AdamW-based online reinforcement learning; and non-dynamic Lanchester evaluation function (L), Simple evaluation function (S), and SimpleSqrt evaluation function (SQ) applied to IDABCD, IDRTMinimax, and Portfolio AI methods.

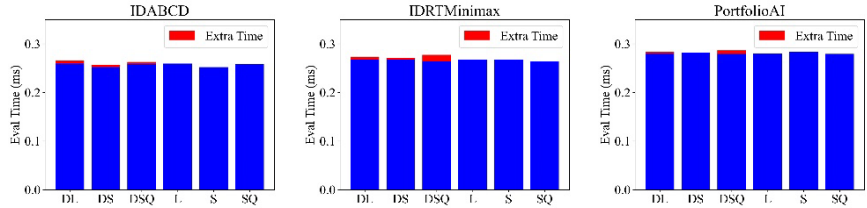

**Fig 9.** Computational efficiency in M3 for Lanchester evaluation function (DL), Simple evaluation function (DS), and SimpleSqrt evaluation function (DSQ) with dynamic weight adjustment using AdamW-based online reinforcement learning; and non-dynamic Lanchester evaluation function (L), Simple evaluation function (S), and SimpleSqrt evaluation function (SQ) applied to IDABCD, IDRTMinimax, and Portfolio AI methods.

## 4    Discussion

Our study presents a novel approach to enhance evaluation functions in real-time strategy (RTS) games by employing an online reinforcement learning-based dynamic weight adjustment mechanism. This method adapts traditional static evaluation functions to the dynamically changing battlefield conditions inherent in RTS games.

The results from our round-robin tournament experiments demonstrate that our proposed dynamic weight adjustment method significantly improves the performance of traditional evaluation functions, including the Lanchester combat model, the Simple evaluation function, and the Simple Sqrt evaluation function. We also found that the dynamic weight adjustment method's effectiveness on evaluation functions varies with algorithm type, map size, and the evaluation function's characteristics. For IDABCD, DSQ and DL saw significant improvements across all map sizes, while DS showed limited improvement on the smaller M1 map due to higher randomness. In IDRTMinimax, the improvements in DS and DSQ became more apparent as map size increased. In the Portfolio method, DS and DSQ consistently improved performance regardless of map size, whereas DL's improvement was minimal. The importance of the type of planning algorithm and map underscore the potential need to consider algorithm specificity, map sizes, and their mutual relationship in dynamic evaluation methods.

One of the major advantages of our approach is its computational efficiency benefited from the inherent advantage of online reinforcement learning. Despite the additional computational steps introduced by the AdamW optimizer, the overall increase in computational time was minimal, averaging less than 6% across all tested evaluation functions and planning algorithms. This ensures that the benefits of improved evaluation accuracy and adaptability do not come at the cost of significant performance overhead, making our method practical for real-time application in RTS games.

The findings align with previous research that highlights the limitations of static evaluation functions and the need for more adaptive approaches. Traditional methods, such as logistic regression-based adjustments or CNN-based models, have shown limited adaptability to dynamic game states and often require extensive computational resources or large datasets for training. Our method leverages online reinforcement learning, specifically the AdamW optimizer, to dynamically adjust evaluation function weights in real-time, thus overcoming the rigidity and computational resource-intensive nature of previous approaches. The practical implications of our study are significant for the development of AI in RTS games. By improving the accuracy and responsiveness of evaluation functions, our method enhances the strategic decision-making capabilities of game AI, leading to more challenging and engaging gameplay for human players. Additionally, the minimal computational overhead ensures that these improvements can be integrated into existing game frameworks without the need for substantial hardware upgrades.

Overall, our dynamic weight adjustment method effectively addresses the limitations of static evaluation functions by enabling real-time adaptability to evolving battlefield conditions, thereby enhancing the accuracy and reliability of AI decision-making. The significant performance gains observed in the DSQ and DL across all map sizes and planning algorithms underscore the effectiveness of our approach in complex and dynamic environments. Additionally, we have modestly acknowledged the scope of our comparisons, noting that while we have demonstrated significant improvements over traditional methods, a broader comparison with all state-of-the-art evaluation techniques remains a potential avenue for future research. This acknowledgment highlights the preliminary nature of our comparative analysis and the need for further studies to comprehensively benchmark our method against the latest advancements in RTS evaluation methodologies.

While our study provides promising results, there are some limitations that warrant further investigation. The performance of the dynamic weight adjustment method varied across different map sizes and planning algorithms, suggesting that further tuning and optimization may be needed to achieve consistent improvements across all scenarios. Additionally, the reliance on the AdamW optimizer, while effective, may not be the only suitable choice for online reinforcement learning in this context. Future research could explore alternative optimization algorithms and their impact on evaluation function performance. Moreover, the integration of more complex features and multi-dimensional input into the evaluation functions could further enhance their accuracy and adaptability. Investigating the application of our method to other experimental platforms and even real-world dynamic systems could also provide valuable insights and broaden the applicability of our approach. Future work may involve integrating our dynamic weight adjustment mechanism with more sophisticated evaluation functions, such as those based on deep learning, to explore synergistic effects and further enhance performance.

# 5 Conclusions

In conclusion, this study introduces a dynamic weight adjustment method for evaluation functions in real-time strategy (RTS) games using an online reinforcement learning approach enhanced by the AdamW optimizer. Our results show that this method significantly improves the adaptability and performance of traditional evaluation functions, particularly in dynamic and unpredictable RTS environments. The effectiveness varied with algorithm type and map size; IDABCD showed significant improvements in DSQ and DL across all map sizes, while DS improved less on smaller maps like M1. In IDRTMinimax, the improvements in DS and DSQ became more pronounced with larger map sizes, whereas Portfolio consistently benefited from DS and DSQ regardless of map size. Despite the additional computational steps, the method maintained high efficiency, with minimal added computational time averaging less than 6%. This dynamic weight adjustment method offers a robust, efficient approach for improving evaluation functions, paving the way for advancements in adaptive AI systems in dynamic environments.

**Author Contributions:** Conceptualization, Weilong Yang; methodology, Yanqing Ye.; software, Jie Zhang; writing—original draft preparation, Jie Zhang.; writing—review and editing, Xunyun Liu. All authors have read and agreed to the published version of the manuscript.

**Funding:** This work was financially supported by the National Natural Science Foundation of China under Grant No.62103438.

**Institutional Review Board Statement:** Not applicable.

**Informed Consent Statement:** Not applicable.

**Data Availability Statement:** The microRTS used in the paper is freely distributed and can be downloaded at https://github.com/Farama-Foundation/MicroRTS (accessed on 7 May 2024).

**Acknowledgments:** In this section, you can acknowledge any support given which is not covered by the author contribution or funding sections. This may include administrative and technical support, or donations in kind (e.g., materials used for experiments).

**Conflicts of Interest:** The authors declare no conflicts of interest.